\pdfoutput=1
\documentclass[11pt]{article}

\usepackage{naacl2021}

\usepackage{times}
\usepackage{latexsym}

\usepackage{graphicx}
\usepackage{amsmath}
\DeclareMathOperator*{\argmin}{arg\,min}

\usepackage{mathtools}

\DeclarePairedDelimiter\abs{\lvert}{\rvert}%
\makeatletter
\let\oldabs\abs
\def\abs{\@ifstar{\oldabs}{\oldabs*}}

\usepackage[T1]{fontenc}
\usepackage[utf8]{inputenc}

\usepackage{microtype}

\title{Finding Pragmatic Differences Between Disciplines}

\author{Lee Kezar \\
  University of Southern California \\
  Information Sciences Institute \\
  \texttt{lkezar@isi.edu} \\\And
  Jay Pujara \\
  University of Southern California \\
  Information Sciences Institute \\
  \texttt{jpujara@isi.edu} \\}

\begin{document}
\maketitle
\begin{abstract}
Scholarly documents have a great degree of variation, both in terms of content (semantics) and structure (pragmatics). Prior work in scholarly document understanding emphasizes semantics through document summarization and corpus topic modeling but tends to omit pragmatics such as document organization and flow. Using a corpus of scholarly documents across 19 disciplines and state-of-the-art language modeling techniques, we learn a fixed set of domain-agnostic descriptors for document sections and ``retrofit'' the corpus to these descriptors (also referred to as ``normalization''). Then, we analyze the position and ordering of these descriptors across documents to understand the relationship between discipline and structure. We report within-discipline structural archetypes, variability, and between-discipline comparisons, supporting the hypothesis that scholarly communities, despite their size, diversity, and breadth, share similar avenues for expressing their work. Our findings lay the foundation for future work in assessing research quality, domain style transfer, and further pragmatic analysis.

\end{abstract}

\section{Introduction}

Disciplines such as art, physics, and political science contain a wide array of ideas, from specific hypotheses to wide-reaching theories. In scholarly research, authors are faced with the challenge of clearly articulating a set of those ideas and relating them to each other, with the ultimate goal of expanding our collective knowledge. In order to understand this work, human readers situate meaning in context \citep{Garten:19}. Similarly, methods for scholarly document processing (SDP) have semantic and pragmatic orientations.

The semantic orientation seeks to understand and evaluate the ideas themselves through information extraction \citep{singh-etal-2016-ocr}, summarization \citep{chandrasekaran-etal-2020-overview-insights}, automatic fact-checking \citep{sathe-etal-2020-automated}, etc. The pragmatic orientation, on the other hand, seeks to understand the context around those ideas through rhetorical and style analysis \citep{august-etal-2020-writing}, corpus topic modeling \citep{paul-girju-2009-topic}, quality prediction \citep{maillette-de-buy-wenniger-etal-2020-structure}, etc. Although both orientations are essential for understanding, the pragmatics of disciplinary writing are very weakly understood.

In this paper, we investigate the structures of disciplinary writing. We claim that a ``structural archetype'' (defined in Section \ref{methods})  can succinctly capture how a community of authors choose to organize their ideas for maximum comprehension and persuasion. Analogous to how syntactic analysis deepens our understanding of a given sentence and document structure analysis deepens our understanding of a given document, structural archetypes, we argue, deepen our understanding of domains themselves.

In order to perform this analysis, we classify sections according to their pragmatic intent. We contribute a data-driven method for deriving the types of pragmatic intent, called a ``structural vocabulary'', alongside a robust method for this classification. Then, we apply these methods to 19k scholarly documents and analyze the resulting structures.

\section{Related Work}

We draw from two areas of related work in SDP: interdisciplinary analysis and rhetorical structure prediction.

In interdisciplinary analysis, we are interested in comparing different disciplines, whether by topic modeling between select corpora/disciplines \citep{paul-girju-2009-topic} or by domain-agnostic language modeling \citep{wang-etal-2020-meta}. These comparisons are more than simply interesting; they allow for models that can adapt to different disciplines, helping the generalizability for downstream tasks like information extraction and summarization.

In rhetorical structure prediction, we are interested in the process of implicature, whether by describing textual patterns in an unsupervised way \citep{o-seaghdha-teufel-2014-unsupervised} or by classifying text as having a particular strategy like ``statistics'' \citep{al-khatib-etal-2017-patterns} or ``analogy'' \citep{august-etal-2020-writing}. These works descend from argumentative zoning \citep{10.1162/coli_a_00364} and the closely related rhetorical structure theory \citep{Mann:88}, which argue that many rhetorical strategies can be described in terms of \textit{units} and their relations. These works are motivated by downstream applications such as predicting the popularity of a topic \citep{prabhakaran-etal-2016-predicting} and classifying the quality of a paper \citep{maillette-de-buy-wenniger-etal-2020-structure}.

Most similar to our work is \citet{arnold-etal-2019-sector}. Here, the authors provide a method of describing Wikipedia articles as a series of section-like topics (e.g. \texttt{disease.symptom}) by clustering section headings into topics and then labeling words and sentences with these topics. We build on this work by using domain-agnostic descriptors instead of domain-specific ones and by comparing structures across disciplines.

\section{Methods} \label{methods}

In this section, we define structural archetypes (\ref{archetypes}) and methods for classifying pragmatic intent through a structural vocabulary (\ref{vocabulary}).

\subsection{Structural Archetypes} \label{archetypes}
We coin the term ``structural archetype'' to focus and operationalize our pragmatic analysis. Here, a ``structure'' is defined as \textit{a sequence of domain-agnostic indicators of pragmatic intent}, while an ``archetype'' refers to \textit{a strong pattern across documents}. In the following paragraphs, we discuss the components of this concept in depth.

\paragraph{Pragmatic Intent} In contrast to verifiable propositions, ``indicators of pragmatic intent'' refer to instances of meta-discourse, comments on the document itself \citep{IFANTIDOU20051325}. There are many examples, including background (comments on what the reader needs in order to understand the content), discussions (comments on how results should be interpreted), and summaries (comments on what is important). These indicators of pragmatic intent serve the critical role of helping readers ``digest'' material; without them, scholarly documents would only contain isolated facts.

We note that the boundary between pragmatic intent and argumentative zones \citep{10.1162/coli_a_00364} is not clear. Some argumentative zones are more suitable for the sentence- and paragraph-level (e.g. ``own claim'' vs. ``background claim'') while others are interpretative (e.g. ``challenge''). This work does not attempt to draw this boundary, and the reader might find overlap between argumentative zoning work and our section types.

\paragraph{Sequences} As a sequence, these indicators reflect how the author believes their ideas should best be received in order to remain coherent. For example, many \textit{background} indicators reflects a belief that the framing of the work is very important.

\paragraph{Domain-agnostic archetypes} Finally, the specification that indicators must be domain-agnostic and that the structures should be widely-held are included to allow for cross-disciplinary comparisons.

We found that the most straightforward way to implement structural archetypes is through classifying section headings according to their pragmatic intent. With this comes a few challenges: (1) defining a set of  domain-agnostic indicators, which we refer to as a ``structural vocabulary''; (2) parsing a document to obtain its structure; and (3) finding archetypes from document-level structures. In the proceeding section, we address (1) and (2), and in Section \ref{disc} we address (3).

\subsection{Deriving a Structural Vocabulary} \label{vocabulary}
Although indicators of pragmatic intent can exist on the sentence level, we follow \citet{arnold-etal-2019-sector} and create a small set of types that are loosely related to common section headings (e.g. ``Methods''). We call this set a ``structural vocabulary'' because it functions in an analogous way to a vocabulary of words; any document can be described as a sequence of items that are taken from this vocabulary. There are three properties that the types should satisfy:
\begin{enumerate}
    \item[A.] \textbf{domain independence}: types should be used by different disciplines
    \item[B.] \textbf{high coverage}: unlabeled instances should be able to be classified as a particular type.
    \item[C.] \textbf{internal consistency}: types should accurately reflect their instances

\end{enumerate}

\paragraph{Domain Independence} As pointed out by \citet{arnold-etal-2019-sector}, there exists a ``vocabulary mismatch problem'' where different disciplines talk about their work in different ways. Indeed, 62\% of the sampled headings only appear once and are not good choices for section types. On the other hand, the most frequent headings are a much better choice, especially those that appear in all domains. After merging a few popular variations among the top 20 section headings (e.g. \textit{conclusion} and \textit{summary}, \textit{background} and \textit{related work}), we yield the following types\footnote{Although \textit{abstract} is extremely common we found it redundant as a section type as it only exists once per paper and in a predictable location.}: \textit{introduction} (a section which introduces the reader to potentially new concepts; $n=10916$), \textit{methods} (a section which details how a hypothesis will be tested; $n=2116$),
\textit{results} (a section which presents findings of the method; $n=3119$), \textit{discussion} (a section which interprets and summarizes the results; $n=3118$), \textit{conclusion} (a section which summarizes the entire paper; $n=7738$), \textit{analysis} (a section which adds additional depth and nuance to the results; $n=951$), and \textit{background} (a section which connects ongoing work to previous related work; $n=800$). Figure \ref{fig:freq_before} contains discipline-level counts.

\paragraph{High Coverage} We can achieve high coverage by classifying any section as one of these section types through language modeling. Specifically, the hidden representation of a neural language model $h(\cdot)$ can act as an embedding of its input. We use the \texttt{[CLS]} tag of SciBERT's hidden layer, selected for its robust representations of scientific literature \citep{beltagy-etal-2019-scibert}.

To classify, we define a distance score $d(\cdot)$ for a section $s$ and a type $T$ as the distance between $h(s)$ and the average embedding across all instances of a type, i.e.

$$ d(s, T) = \abs{ h(s) - \frac{\sum_{t\in T} h(t)}{\|T\|} }$$

Note that since the embedding is a vector, addition and division are elementwise. Then, we compute the distance for all types in the vocabulary $V$ and select the minimum, i.e.

$$ s_{type} = \argmin_{T \in V} (d(s, T))$$

\paragraph{Internal Consistency} Some sections do not adequately fit any section type, so nearest-neighbor classification will result in very inconsistent clusters. We address this problem by imposing a threshold on the maximum distance for $d(\cdot)$. Further, since the types have unequal variance (that is, the ground truth for some types are more consistent than other types), we define a type-specific threshold as half of the distance from the center of $T$ to the furthest member of $T$, i.e.

$$ \textrm{threshold}_T = 0.5 \cdot \max_{t \in T}(d(t, T))$$

The weight of 0.5 was found to remove outliers appropriately an maximize retrofitting performance (Section \ref{retro_perf}).

We also note that some headings, especially brief ones, leave much room for interpretation and make retrofitting challenging. We address this problem by concatenating tokens of each section's heading and body, up to 25 tokens, as input to the language model. This ensures that brief headings contain enough information to make an accurate representation without including too many details from the body text.

\section{Results and Discussion} \label{disc}

\subsection{Data}
We use the Semantic Scholar Open Research Corpus (S2ORC) for all analysis \citep{lo-etal-2020-s2orc}. This corpus, which is freely available, contains approximately 7.5M PDF-parsed documents from 19 disciplines, including natural sciences, social sciences, arts, and humanities. For our experiments, we randomly sample 1k documents for each discipline, yielding a total of 19k documents. 

\subsection{Retrofitting Performance} \label{retro_perf}
Retrofitting (or normalizing) section headers refers to re-labeling sections with the structural vocabulary. We evaluate retrofitting performance by manually tagging 30 of each section type and comparing the true labels to the predicted values. Our method yields an average F1 performance of 0.76. The breakdown per section type, shown in Table \ref{tab:retrofit_perf}, reveals that \textit{conclusion}, \textit{background}, and \textit{analysis} sections were the most difficult to predict. We attribute this to a lack of textual clues in the heading and body, and also a semantic overlap with \textit{introduction} sections. Future work can improve the classifier with more nuanced signals, such as position, length, number of references, etc.

\begin{table}[h]
    \centering
    \begin{tabular}{|c|c|c|c|}
        \hline
        Type & Precision & Recall & F1\\
        \hline
        introduction & 0.77 & 0.97 & 0.85\\
        \hline
        conclusion & 0.67 & 0.72 & 0.69\\
        \hline
        discussion & 0.88 & 0.88 & 0.88\\
        \hline
        results & 0.80 & 0.85 & 0.83\\
        \hline
        methods & 0.83 & 0.91 & 0.87\\
        \hline
        background & 0.63 & 0.77 & 0.69\\
        \hline
        analysis & 0.50 & 0.61 & 0.55\\
        \hline
        \textbf{overall} & \textbf{0.72} & \textbf{0.88} & \textbf{0.76} \\
        \hline
    \end{tabular}
    \caption{Type-level and overall performance for section type retrofitting.}
    \label{tab:retrofit_perf}
\end{table}

\begin{figure*}[t]
    \centering
    \includegraphics[width=0.45\textwidth]{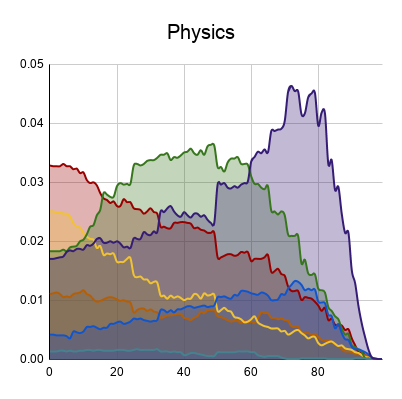} %
    \includegraphics[width=0.45\textwidth]{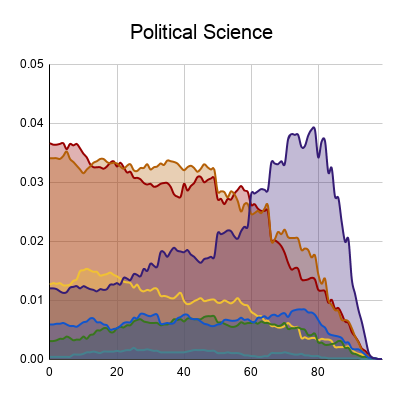}
    \includegraphics[width=\textwidth]{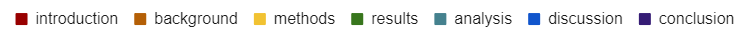}
    \caption{A comparison between the positions (normalized by document length; $x$ axis) and frequencies ($y$ axis) of section types in Physics and Political Science. Comparable distributions of \textit{introduction}, \textit{methods}, \textit{analysis}, \textit{discussion}, and \textit{conclusion}, but different distributions of background and results.}
    \label{fig:phys_poli}
\end{figure*}

\subsection{Analyzing Position with Aggregate Frequency}
A simple yet expressive way of showing the structural archetypes of a discipline is to consider the frequency of a particular type at any point in the article (normalized by length). This analysis reveals general trends throughout a discipline's documents, such as where a section type is most frequent or where there is homogeneity.

To illustrate the practicality of this analysis, consider the hypothesis that Physics articles are more empirically-motivated while Political Science articles are more conceptually-motivated, i.e. that they are on opposing ends of the \textit{concrete} versus \textit{abstract} spectrum. We operationalize this by claiming that Physics articles have more \textit{methods}, \textit{results}, and \textit{analysis} sections than Political Science. Figure \ref{fig:phys_poli} shows the difference between Physics and Political Science at each point in the article. It reveals that not only do Physics articles contain more \textit{methods} and \textit{results}, but also that Physics articles introduce \textit{methods} earlier than Political Science, and that both contain the same amount of \textit{analysis} sections.

\subsection{Analyzing Ordering with State Transitions}
A more structural analysis of a discipline is to look at the frequency of sequence fragments through computing transition probabilities. As a second example, suppose we have a more nuanced hypothesis: that Psychology papers tend to separate claims and evaluate them sequentially (\textit{methods}, \textit{results}, \textit{discussion}, repeat) whereas Sociology papers tend to evaluate all claims at once. We can operationalize these hypotheses by calculating the transition probability between section $s_i$ and $s_{i-1}$ conditioned on some discipline.

In Table \ref{tab:transitions}, we see evidence that \textit{methods} sections are more likely to be preceded by \textit{results} sections in Psychology than Sociology, implying a new iteration of a cycle. We might conclude that Psychology papers are more likely to have cyclical experiments, but not that Sociology papers conduct multiple experiments in a linear fashion.

\begin{table}[h!]
    \centering
    \begin{tabular}{ |p{1.3cm} l| c | c |}
    \hline
    \multicolumn{2}{|c|}{Transition Probability} & Psych. & Socio. \\
    \hline
    $P(\textit{method}$ & $\rightarrow \textit{method})$ & $0.31$ & $0.20$ \\
    \hline
    $P(\textit{results}$ & $\rightarrow \textit{results})$ & $0.22$ & $0.23$ \\
    \hline
    $P(\textit{disc}$ & $\rightarrow \textit{disc})$ & $0.16$ & $0.13$ \\
    \hline
    $P(\textit{method}$ & $\rightarrow \textit{results})$ & $0.21$ & $0.10$ \\
    \hline
    $P(\textit{results}$ & $\rightarrow \textit{disc})$ & $0.15$ & $0.16$ \\
    \hline
    $P(\textit{disc}$ & $\rightarrow \textit{method})$ & $0.23$ & $0.13$ \\
    \hline
    \end{tabular}
    \caption{Transition probabilities for methods, results, and discussion in Psychology and Sociology}
    \label{tab:transitions}
\end{table}

\section{Conclusion and Future Work}

In this paper, we have shown a simple method for constructing and comparing structural archetypes across different disciplines. By classifying the pragmatic intent of section headings, we can visualize structural trends across disciplines. In addition to utilizing a more complex classifier, future directions for this work include (1) further distinguishing between subdisciplines (e.g. abnormal psychology vs. developmental psychology) and document type (e.g. technical report vs. article); (2) learning relationships between structures and measures of research quality, such as reproducibility; (3) learning how to convert one structure into another, with the ultimate goal of normalizing them for easier comprehension or better models; (4) deeper investigations into the selection of a structural vocabulary, such as including common argumentative zoning types or adjusting the scale to the sentence-level; and (5) drawing comparisons, such as by clustering, between different documents based strictly on their structure.

\section{Acknowledgements}

This work was funded by the Defense Advanced Research Projects Agency with award W911NF-19-20271. The authors would like to thank the reviewers of this paper for their detailed and constructive feedback, and in particular their ideas for future directions.

\bibliography{anthology,custom}

\begin{thebibliography}{17}
\expandafter\ifx\csname natexlab\endcsname\relax\def\natexlab#1{#1}\fi

\bibitem[{Al-Khatib et~al.(2017)Al-Khatib, Wachsmuth, Hagen, and
  Stein}]{al-khatib-etal-2017-patterns}
Khalid Al-Khatib, Henning Wachsmuth, Matthias Hagen, and Benno Stein. 2017.
\newblock \href {https://doi.org/10.18653/v1/D17-1141} {Patterns of
  argumentation strategies across topics}.
\newblock In \emph{Proceedings of the 2017 Conference on Empirical Methods in
  Natural Language Processing}, pages 1351--1357, Copenhagen, Denmark.
  Association for Computational Linguistics.

\bibitem[{Arnold et~al.(2019)Arnold, Schneider, Cudr{\'e}-Mauroux, Gers, and
  L{\"o}ser}]{arnold-etal-2019-sector}
Sebastian Arnold, Rudolf Schneider, Philippe Cudr{\'e}-Mauroux, Felix~A. Gers,
  and Alexander L{\"o}ser. 2019.
\newblock \href {https://doi.org/10.1162/tacl_a_00261} {{SECTOR}: A neural
  model for coherent topic segmentation and classification}.
\newblock \emph{Transactions of the Association for Computational Linguistics},
  7:169--184.

\bibitem[{August et~al.(2020)August, Kim, Reinecke, and
  Smith}]{august-etal-2020-writing}
Tal August, Lauren Kim, Katharina Reinecke, and Noah~A. Smith. 2020.
\newblock \href {https://doi.org/10.18653/v1/2020.emnlp-main.429} {Writing
  strategies for science communication: Data and computational analysis}.
\newblock In \emph{Proceedings of the 2020 Conference on Empirical Methods in
  Natural Language Processing (EMNLP)}, pages 5327--5344, Online. Association
  for Computational Linguistics.

\bibitem[{Beltagy et~al.(2019)Beltagy, Lo, and
  Cohan}]{beltagy-etal-2019-scibert}
Iz~Beltagy, Kyle Lo, and Arman Cohan. 2019.
\newblock \href {https://doi.org/10.18653/v1/D19-1371} {{S}ci{BERT}: A
  pretrained language model for scientific text}.
\newblock In \emph{Proceedings of the 2019 Conference on Empirical Methods in
  Natural Language Processing and the 9th International Joint Conference on
  Natural Language Processing (EMNLP-IJCNLP)}, pages 3615--3620, Hong Kong,
  China. Association for Computational Linguistics.

\bibitem[{Chandrasekaran et~al.(2020)Chandrasekaran, Feigenblat, Hovy,
  Ravichander, Shmueli-Scheuer, and
  de~Waard}]{chandrasekaran-etal-2020-overview-insights}
Muthu~Kumar Chandrasekaran, Guy Feigenblat, Eduard Hovy, Abhilasha Ravichander,
  Michal Shmueli-Scheuer, and Anita de~Waard. 2020.
\newblock \href {https://doi.org/10.18653/v1/2020.sdp-1.24} {Overview and
  insights from the shared tasks at scholarly document processing 2020:
  {CL}-{S}ci{S}umm, {L}ay{S}umm and {L}ong{S}umm}.
\newblock In \emph{Proceedings of the First Workshop on Scholarly Document
  Processing}, pages 214--224, Online. Association for Computational
  Linguistics.

\bibitem[{Ifantidou(2005)}]{IFANTIDOU20051325}
Elly Ifantidou. 2005.
\newblock \href {https://doi.org/https://doi.org/10.1016/j.pragma.2004.11.006}
  {The semantics and pragmatics of metadiscourse}.
\newblock \emph{Journal of Pragmatics}, 37(9):1325--1353.
\newblock Focus-on Issue: Discourse and Metadiscourse.

\bibitem[{Justin~Garten and Deghani(2019)}]{Garten:19}
Kenji~Sagae Justin~Garten, Brendan~Kennedy and Morteza Deghani. 2019.
\newblock \href {https://doi.org/10.3758/s13428-019-01200-w} {Measuring the
  importance of context when modeling language comprehension}.
\newblock \emph{Behavioral Research Methods}, 51:480--492.

\bibitem[{Lawrence and Reed(2020)}]{10.1162/coli_a_00364}
John Lawrence and Chris Reed. 2020.
\newblock \href {https://doi.org/10.1162/coli_a_00364} {{Argument Mining: A
  Survey}}.
\newblock \emph{Computational Linguistics}, 45(4):765--818.

\bibitem[{Lo et~al.(2020)Lo, Wang, Neumann, Kinney, and
  Weld}]{lo-etal-2020-s2orc}
Kyle Lo, Lucy~Lu Wang, Mark Neumann, Rodney Kinney, and Daniel Weld. 2020.
\newblock \href {https://doi.org/10.18653/v1/2020.acl-main.447} {{S}2{ORC}: The
  semantic scholar open research corpus}.
\newblock In \emph{Proceedings of the 58th Annual Meeting of the Association
  for Computational Linguistics}, pages 4969--4983, Online. Association for
  Computational Linguistics.

\bibitem[{Maillette~de Buy~Wenniger et~al.(2020)Maillette~de Buy~Wenniger, van
  Dongen, Aedmaa, Kruitbosch, Valentijn, and
  Schomaker}]{maillette-de-buy-wenniger-etal-2020-structure}
Gideon Maillette~de Buy~Wenniger, Thomas van Dongen, Eleri Aedmaa, Herbert~Teun
  Kruitbosch, Edwin~A. Valentijn, and Lambert Schomaker. 2020.
\newblock \href {https://doi.org/10.18653/v1/2020.sdp-1.18} {Structure-tags
  improve text classification for scholarly document quality prediction}.
\newblock In \emph{Proceedings of the First Workshop on Scholarly Document
  Processing}, pages 158--167, Online. Association for Computational
  Linguistics.

\bibitem[{Mann and Thompson(1988)}]{Mann:88}
William Mann and Sandra Thompson. 1988.
\newblock \href {https://doi.org/10.1515/text.1.1988.8.3.243} {Rhetorical
  structure theory: Toward a functional theory of text organization}.
\newblock \emph{Text - Interdisciplinary Journal for the Study of Discourse},
  8:243--281.

\bibitem[{{\'O}~S{\'e}aghdha and
  Teufel(2014)}]{o-seaghdha-teufel-2014-unsupervised}
Diarmuid {\'O}~S{\'e}aghdha and Simone Teufel. 2014.
\newblock \href {https://www.aclweb.org/anthology/C14-1002} {Unsupervised
  learning of rhetorical structure with un-topic models}.
\newblock In \emph{Proceedings of {COLING} 2014, the 25th International
  Conference on Computational Linguistics: Technical Papers}, pages 2--13,
  Dublin, Ireland. Dublin City University and Association for Computational
  Linguistics.

\bibitem[{Paul and Girju(2009)}]{paul-girju-2009-topic}
Michael Paul and Roxana Girju. 2009.
\newblock \href {https://www.aclweb.org/anthology/R09-1061} {Topic modeling of
  research fields: An interdisciplinary perspective}.
\newblock In \emph{Proceedings of the International Conference {RANLP}-2009},
  pages 337--342, Borovets, Bulgaria. Association for Computational
  Linguistics.

\bibitem[{Prabhakaran et~al.(2016)Prabhakaran, Hamilton, McFarland, and
  Jurafsky}]{prabhakaran-etal-2016-predicting}
Vinodkumar Prabhakaran, William~L. Hamilton, Dan McFarland, and Dan Jurafsky.
  2016.
\newblock \href {https://doi.org/10.18653/v1/P16-1111} {Predicting the rise and
  fall of scientific topics from trends in their rhetorical framing}.
\newblock In \emph{Proceedings of the 54th Annual Meeting of the Association
  for Computational Linguistics (Volume 1: Long Papers)}, pages 1170--1180,
  Berlin, Germany. Association for Computational Linguistics.

\bibitem[{Sathe et~al.(2020)Sathe, Ather, Le, Perry, and
  Park}]{sathe-etal-2020-automated}
Aalok Sathe, Salar Ather, Tuan~Manh Le, Nathan Perry, and Joonsuk Park. 2020.
\newblock \href {https://www.aclweb.org/anthology/2020.lrec-1.849} {Automated
  fact-checking of claims from {W}ikipedia}.
\newblock In \emph{Proceedings of the 12th Language Resources and Evaluation
  Conference}, pages 6874--6882, Marseille, France. European Language Resources
  Association.

\bibitem[{Singh et~al.(2016)Singh, Barua, Palod, Garg, Satapathy, Bushi, Ayush,
  Sai~Rohith, Gamidi, Goyal, and Mukherjee}]{singh-etal-2016-ocr}
Mayank Singh, Barnopriyo Barua, Priyank Palod, Manvi Garg, Sidhartha Satapathy,
  Samuel Bushi, Kumar Ayush, Krishna Sai~Rohith, Tulasi Gamidi, Pawan Goyal,
  and Animesh Mukherjee. 2016.
\newblock \href {https://www.aclweb.org/anthology/C16-1320} {{OCR}++: A robust
  framework for information extraction from scholarly articles}.
\newblock In \emph{Proceedings of {COLING} 2016, the 26th International
  Conference on Computational Linguistics: Technical Papers}, pages 3390--3400,
  Osaka, Japan. The COLING 2016 Organizing Committee.

\bibitem[{Wang et~al.(2020)Wang, Qiu, Huang, and He}]{wang-etal-2020-meta}
Chengyu Wang, Minghui Qiu, Jun Huang, and Xiaofeng He. 2020.
\newblock \href {https://doi.org/10.18653/v1/2020.emnlp-main.250} {Meta
  fine-tuning neural language models for multi-domain text mining}.
\newblock In \emph{Proceedings of the 2020 Conference on Empirical Methods in
  Natural Language Processing (EMNLP)}, pages 3094--3104, Online. Association
  for Computational Linguistics.

\end{thebibliography}
\bibliographystyle{acl_natbib}

\appendix
\onecolumn
\section{Section Counts Before and After Retrofitting}

\begin{figure}[h]
    \centering
    \includegraphics[width=0.8\textwidth]{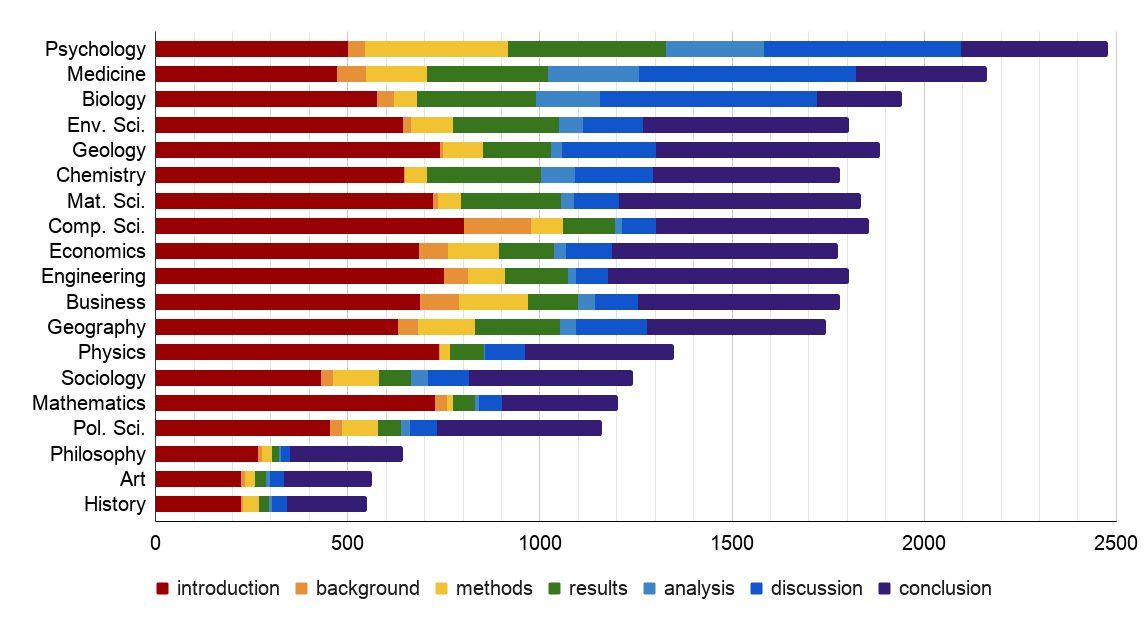}
    \includegraphics[width=0.8\textwidth]{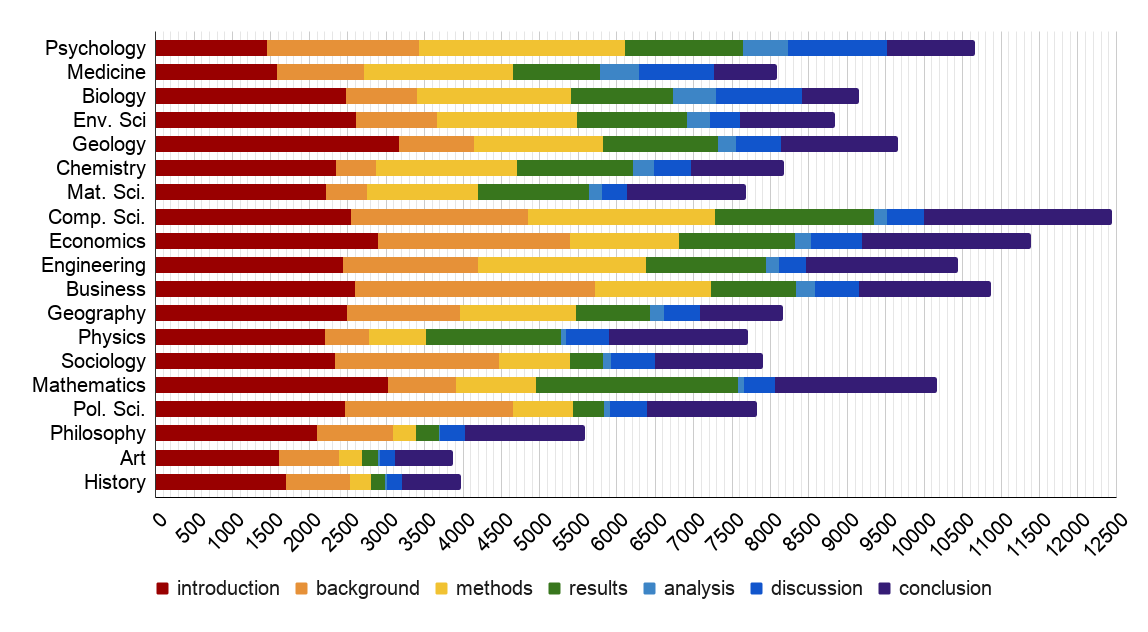}
    \caption{The frequency of the top-7 section headings before (top) and after (bottom) retrofitting. }
    \label{fig:freq_before}
\end{figure}

\section{Aggregate Frequency for Other Disciplines}

\begin{figure}[h]
    \centering
    \includegraphics[width=0.9\textwidth]{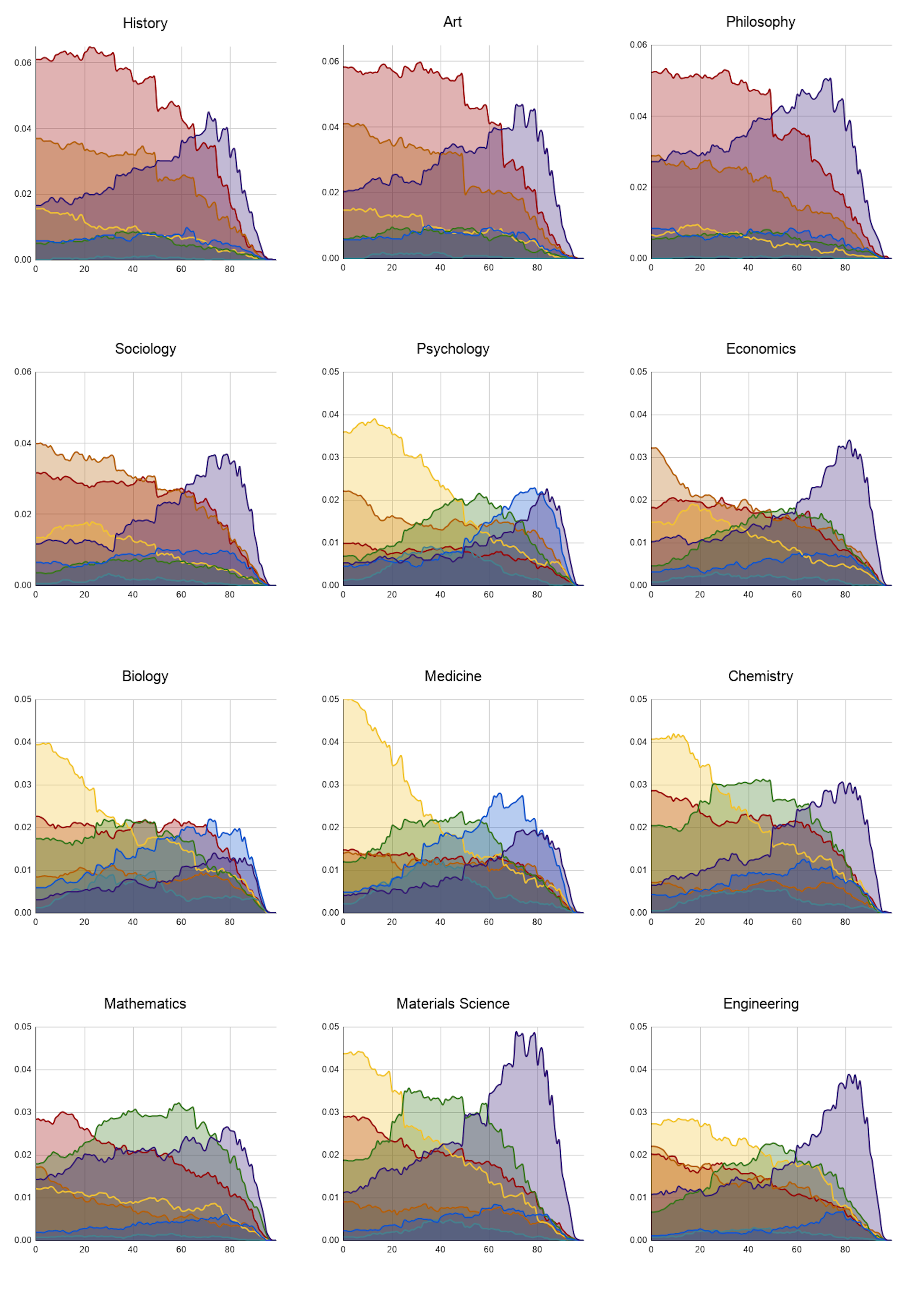}
    \includegraphics[width=0.9\textwidth]{positions/legend.png}
    \caption{Aggregate Frequency for 12 of the 19 disciplines}
    \label{fig:positions}
\end{figure}

\end{document}